\documentclass{article}
\usepackage{ijcai23}

\usepackage{times}
\usepackage{soul}
\usepackage{url}
\usepackage[hidelinks]{hyperref}
\usepackage[utf8]{inputenc}
\usepackage[small]{caption}
\usepackage{graphicx}
\usepackage{amsmath}
\usepackage{amsthm}
\usepackage{booktabs}
\usepackage{algorithm}
\usepackage{algorithmic}
\usepackage[switch]{lineno}
\usepackage{natbib}
\usepackage{tabularx}

\usepackage{url}
\usepackage{comment}

\title{Towards Theory-based Moral AI:\\ Moral AI with Aggregating Models Based on  Normative Ethical Theory}

\author{
Masashi Takeshita$^1$
\and
Rzepka Rafal$^2$\And
Kenji Araki$^2$
\affiliations
$^1$Graduate School of Information Science and Technology, Hokkaido University.\\
$^2$Faculty of Information Science and Technology, Hokkaido University.\\
\emails
\{takeshita.masashi, rzepka, araki\}@ist.hokudai.ac.jp
}

\begin{document}

\maketitle

\begin{abstract}
Moral AI has been studied in the fields of philosophy and artificial intelligence. Although most existing studies are only theoretical, recent developments in AI have made it increasingly necessary to implement AI with morality. On the other hand, humans are under the moral uncertainty of not knowing what is morally right. In this paper, we implement the Maximizing Expected Choiceworthiness (MEC) algorithm, which aggregates outputs of models based on three normative theories of normative ethics to generate the most appropriate output. MEC is a method for making appropriate moral judgments under moral uncertainty. Our experimental results suggest that the output of MEC correlates to some extent with commonsense morality and that MEC can produce equally or more appropriate output than existing methods.
\end{abstract}

\section{Introduction}

Philosophy and artificial intelligence have long considered the creation of \textit{Moral AI} by which we mean artificial intelligence with morality\footnote{We use ``moral'' and ``ethical'' interchangeably.}. In philosophy, theoretical studies have explored under what framework the creation of moral AI is desirable~\citep{wallach-allen2008moral-machines,allen2005artificial-morality}, while the filed of AI is still exploring how to implement such a framework~\citep[e.g.][]{anderson2006medethex,Rzepka2017WhatPS,Jiang-2021-can-machine-learn-morality-delphi}.

There are many reasons why Moral AI is essential. For example, if an AI is implemented in automated driving technology, it will likely make morally wrong decisions if it cannot correctly make moral judgments~\citep{awad2018moral-machine-experiment}.
Similarly, if healthcare workers use AI for decision-making support in the medical field, that AI must be able to make appropriate ethical decisions~\citep{Braune-2021-primer-ethics-ai-decision-support-clinic}.
Furthermore, as AI becomes more accessible and assists or advises us in various aspects of our daily lives, it may lead us in the wrong direction if such AI ignores morality.

In recent years, large language models (LLMs) have been developed, and the implementation of morality has become essential, but most of previous research on moral AIs has been done without implementation~\citep{tolmeijer2020implementations-machine-ethics}.
The social impact of foundation models~\citep{bommasani2021opportunities-risks-foundation-models}, such as the BERT~\citep{devlin-etal-2019-bert} and GPT series~\citep[e.g.][]{brown2020language-few-shot-leaner-gpt-3}, is significant, and it is essential to implement appropriate ethics in these foundational models.
Recently, users have an easy access to output of LLMs, which is known to contain harmful content and discriminatory bias~\citep{gehman-etal-2020-realtoxicityprompts,ganguli2022red-teaming-languagemodel-method-behavior}. Therefore, it has become more important to implement appropriate ethics in LLM. But what ethics should we implement in AI?

\begin{figure}
    \centering
    \includegraphics[width=0.9\linewidth]{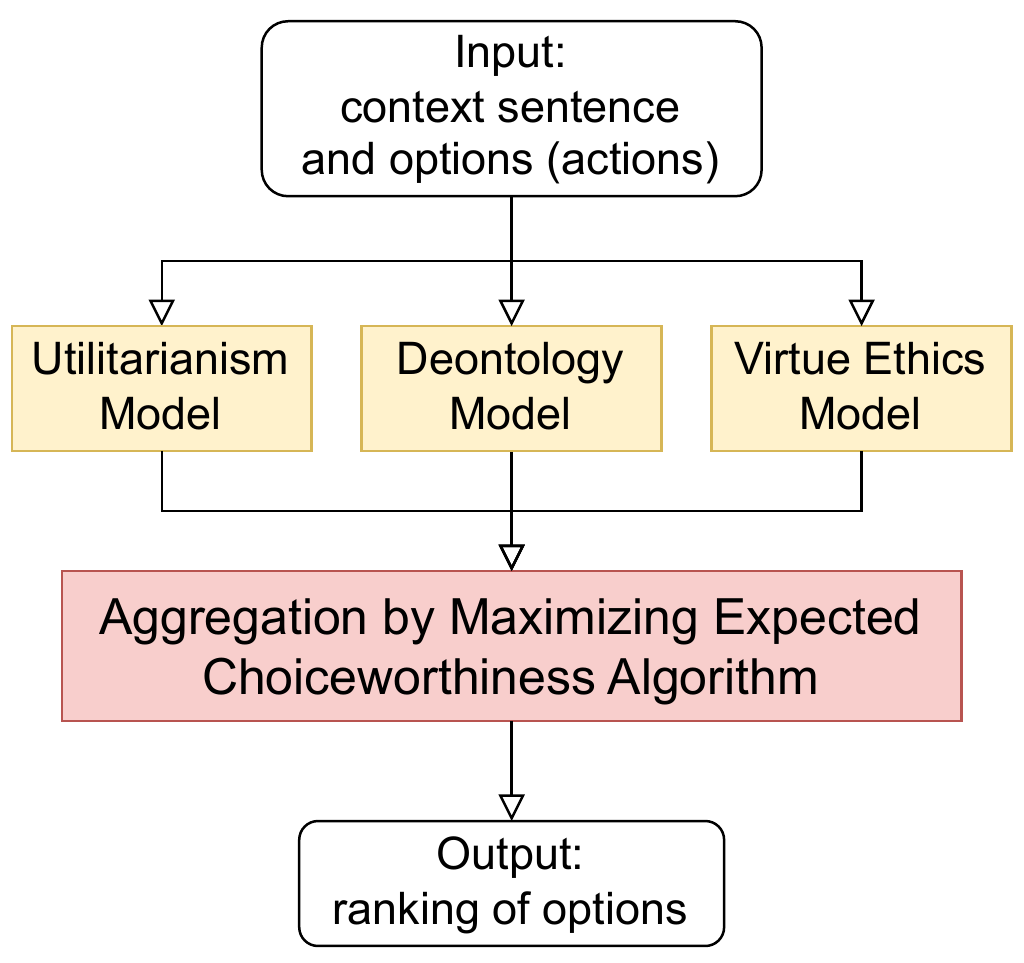}
    \caption{This figure shows the overall picture of the aggregation model in this paper. 
    In this paper, we use a theory-based model based on utilitarianism, deontology, and virtue ethics.
    }
    \label{fig:mec-algorithm}
\end{figure}

To answer this question, we create several models based on normative ethical theories studied in normative ethics and implement an algorithm to aggregate the output of these models (Figure \ref{fig:mec-algorithm}).
We call the algorithm \textit{Maximizing Expected Choiceworthiness} (MEC) algorithm~\citep{macaskill-bykvist-ord2020moral-uncertainty}.
Existing research has created models and datasets solely based on commonsense morality~\citep[e.g.][]{Jiang-2021-can-machine-learn-morality-delphi}. 
However, relying directly on commonsense morality may be inappropriate if it is incorrect. 
Therefore, we use findings of normative ethics and implement Moral AI based on the normative theories studied there.
It is possible to create morally appropriate AI by not relying on commonsense morality as it is, \textit{but} by referring to various normative
theories.
Although this idea has already been proposed~\citep{bogosian2017implementation,macaskill-bykvist-ord2020moral-uncertainty}, no implementation and evaluation experiments have been conducted. Thus, we improve, implement, and evaluate this idea.

The structure of this paper is as follows.
In Section 2 we describe existing research on implementations of morality in AI, and on moral uncertainty, a central concept in this research.
In Section 3 we describe Maximizing Expected Choiceworthiness (MEC) algorithm.
Section 4 explains why MEC is desirable in the implementation of Moral AI and Section 5 describes the experiments. 
Section 6 describes the experimental results, which are discussed in Section 7.

\section{Related Works}

\subsection{MoralQA and Moral AI}
AI researchers have been trying to implement AIs capable of making moral judgments (we will call them \textit{Moral AI(s)} in this paper) in various ways.
In one of early examples, \citet{anderson2006medethex} have proposed MedEth, which learns by inductive logic programming based on principles proposed by ethicists, and advises the user. MedEth is limited to situations related to medical ethics, but its generalized version, GenEth~\citep{anderson-anderson-2018-geneth}, has also been proposed.

Since the advent of deep learning, researchers have studied MoralQA, which is to study whether AI can predict human answers to questions about morality~\citep{Jin-2022-when-to-make-exceptions-moral-exceptqa}.
There are two types of datasets used in the MoralQA task: ones based on ethical theories and ones that not.
The ETHICS dataset \citep{hendrycks2021aligning-AI-human-values} is created based on ethical theories. It consists of five datasets: justice, utilitarianism, duty theory, virtue, and commonsense morality. Except for commonsense morality, four datasets were created based on their respective theories and concepts.
\citet{Jin-2022-when-to-make-exceptions-moral-exceptqa} also developed a dataset called MoralExceptQA to assess understanding of when it is acceptable to break the rules based on contractualism.

One example of a QA dataset that is not explicitly based on ethical theory is Social Chemistry 101~\citep{forbes-etal-2020-social-chemistry-101}. \citet{forbes-etal-2020-social-chemistry-101} asked crowdworkers a) to create situation-related rules of thumb (RoTs) and to annotate b) which category of morality the RoTs belong to and c) the moral evaluation of the actions in the situation.
Another example of this type of dataset is the Moral Stories dataset~\citep{emelin-etal-2021-moral-stories} which was created based on Social Chemistry 101. Moral Stories consists of norms, situations, related intentions and actions, and consequences of actions.
\citet{Jiang-2021-can-machine-learn-morality-delphi} have re-edited various commonsense morality datasets, including Social Chemistry 101 and MoralStories, to compile approximately 1.7 million data from The Commonsense Norm Bank.

As a model for solving MoralQA tasks, \citep{Jiang-2021-can-machine-learn-morality-delphi} created Delphi, which can answer moral questions using the Commonsense Norm Bank.
Delphi is a model based on UNICORN~\citep{lourie2021unicorn-rainbow}, a model built on T5~\citep{Raffel-2020-t5}, and further trained with Commonsense Norm Bank.
UNICORN is a model trained using RAINBOW~\citep{lourie2021unicorn-rainbow}, a collection of CommonsenseQA datasets.
Therefore, Delphi is a model trained on the commonsense morality dataset and not on a dataset that explicitly references ethical theory.

\citet{Jiang-2021-can-machine-learn-morality-delphi} suggested two approaches to the creation of Moral AI, top-down and bottom-up \citep[cf.][]{wallach-allen2008moral-machines}, and stated that Delphi is based on a bottom-up approach. The top-down approach is to create Moral AI by referring to moral theories and rules, while the bottom-up approach is to create Moral AI based on data such as people's intuition. 
As we have seen, most of the existing research is bottom-up (the exception is \citet{anderson2006medethex}).
However, there are problems with using a bottom-up approach alone, such as being conservative because it is based on current commonsense morality and cannot correct commonsense morality when it is wrong.
As \citet{Jiang-2021-can-machine-learn-morality-delphi} correctly point out, top-down and bottom-up approaches must be mutually influential.
Because we train AI models based on each moral theory, our study belongs to the top-down approach. 
Therefore, our research is meant to complement the existing bottom-up approach.

\subsection{Moral Uncertainty}
Moral uncertainty is ``uncertainty that stems not from uncertainty about descriptive matters, but about moral or evaluative matters.'' \citep[p.1]{macaskill-bykvist-ord2020moral-uncertainty}.
Moral uncertainty arises not from uncertainty about descriptive questions but from evaluative questions.
For example, in normative ethics, various theories have been proposed, such as utilitarianism and deontology, and no one yet knows which is the correct theory. Which theory is favored may still be open even if all the descriptive problems are solved.

We must make moral decisions without knowing which theory is correct.
\citet{bogosian2017implementation} points out two problems with this moral uncertainty.
First, moral disagreement makes cooperation among engineers, policymakers, and philosophers difficult. In some cases, it can become simply an ideological conflict.
Second, if there is wide disagreement about normative ethical theories, making moral judgments based on a single theory in the presence of diverse positions is, statistically speaking, unlikely to be the right decision.
Because of the second problem, some philosophers avoid top-down approaches. However, due to the existence of the first problem and moral disagreement, bottom-up approaches are also unlikely to resolve moral dilemmas or politically divisive issues.
Therefore, implementing Moral AI based on a single principle is undesirable, and relying solely on a bottom-up approach is problematic.

Some researchers \citep{macaskill-bykvist-ord2020moral-uncertainty, bogosian2017implementation} proposed \textit{Maximizing Expected Choiceworthiness (MEC)} algorithm as a desirable decision-making approach in situations of moral uncertainty. We will explain this idea below, particularly regarding AI implementations.

\section{Maximizing Expected Choiceworthiness As a Solution to Moral Uncertainty}

Maximizing Expected Choiceworthiness (MEC) is one of the solutions to decision-making problems in moral uncertain situations. 
This section describes this framework following \citet[sec. 5]{bogosian2017implementation}'s explanation \citep[cf.][]{macaskill-bykvist-ord2020moral-uncertainty}.

\subsection{Preliminary Definition}
An AI chooses an action in  a decision situation $<S, t, \mathcal{A}, T, C>$ where $S$ is the decision maker, $t$ is the time, and $\mathcal{A}$ is the set of possible actions (options) to take.
$T$ is the set of normative theories under consideration. A theory $T_i$ is a function of decision that produces a cardinal or ordinal choiceworthiness score for actions $CW_i(A)$ for all actions $a\in \mathcal{A}$. 
$C(T_i)$ is a credence function that assigns values in $[0, 1]$ to every $T_i \in T$. A metanormative theory is a function of decision-situations that produces an ordering of the actions in $\mathcal{A}$ regarding their appropriateness.

$T$ includes three kinds of moral theories: (1) theories that assign a cardinal ranking to options, and these rankings are comparable, (2) theories that assign a cardinal ranking to options, and these rankings are incomparable and (3) theories that assign ordinal rankings to options. For example, typically, utilitarianism is a cardinal theory, and deontology is an ordinal theory.

In the case of current AI models, because they can output the probability of a given label in the range of $[0, 1]$, we can interpret all outputs as cardinal values. However, probability is not a choiceworthiness itself, so we treat the values assigned by AI models based on ordinal theories as ordinal scale values.

\subsection{Calculating Choiceworthiness}
Maximizing Expected Choiceworthiness (MEC) consists of four steps of calculating choiceworthiness scores.
\paragraph{Step 1: Merging commensurable cardinal theories} Given a set of $k$ intertheoretically comparable cardinal theories and $k$ sets of actions assigned choiceworthiness scores by each theory, these theories are merged into a single theory $\mathcal{K}$:
\begin{align}
    C\left( {T_{{\mathcal{K}}} } \right) &= \mathop \sum \limits_{i = 1}^{k} C(T_{i} )\\
    CW_{{\mathcal{K}}} \left( A \right) &= \frac{{\mathop \sum \nolimits_{i = 1}^{k} CW_{i} \left( A \right)C\left( {T_{i} } \right)  }}{{\mathop \sum \nolimits_{i = 1}^{k} C\left( {T_{i} } \right)}}
\end{align}

\paragraph{Step 2: Assigning choiceworthiness scores by ordinal theories}
A modified Borda scoring rule is used to generate scores for each action based on the ranking of actions which is provided on each ordinal theory $o$, considering ties. These scores are represented as $CW^B$. The score of an action is determined by the difference between the number of actions inferior to it and the number of actions superior to it.
\begin{equation}
  \begin{split} 
    CW_{o}^{B}& \left( A \right) = \\
    &\left| {a \in {\mathcal{A} }:CW_{o} \left( a \right) < CW_{o} \left( A \right)} \right|\\
    & - \left| {a \in {\mathcal{A} }:CW_{o} \left( a \right)} \right > CW_{o} \left( A \right)|
  \end{split}
\end{equation}
An AI model outputs the probability of given labels. Here, we have two ways of treating this probability. 
First, we can treat this probability as a cardinal value. For example, if an AI model outputs 0.8 for an action $a_1$ and 0.6 for $a_2$ for the probability that the label is ``1'' (e.g., wrong), we treat $a_1$ as having a higher choiceworthiness score than $a_2$.
Second, we can use threshold and treat model outputs as ordinal values. For instance, if an AI model outputs 0.8 for an action $a_1$ and 0.6 for $a_2$ and we set 0.5 as a threshold, we treat both $a_1$ and $a_2$ as ``1'' (e.g., wrong) equally.

\paragraph{Step 3: Normalization}
To equalize the value of voting for each value system, all choiceworthiness scores $CW_i(A)$ are divided by their respective standard deviations:
\begin{align}
    CW_{{\mathcal{K}}}^{N} (A) &= \frac{{CW_{{\mathcal{K}}} (A)}}{{\sigma \left( {CW_{{\mathcal{K}}} ({\mathcal{G}})} \right)}}\\
    CW_{o}^{N} ( A ) &= \frac{{CW_{o}^{B} (A)}}{{{\sigma}\left( {CW_{o}^{B} \left( {\mathcal{G}} \right)} \right)}}\\
    CW_{p}^{N} \left( A \right) &= \frac{{  CW_{p} \left( A \right)}}{{{\sigma}\left( {CW_{p} \left( {\mathcal{G}} \right)} \right)}}
\end{align}
where theories $p$ means intertheoretically incomparable cardinal theories, $\mathcal{G}$ is a representative set of actions (the ``general set''), which actions may not be included $\mathcal{A}$.
\citet{bogosian2017implementation} stated the choiceworthiness score should be divided by standard deviations of ${CW_{o}^{B} \left( {\mathcal{G}} \right)}$, because we should think about whether each considered action ``is comparatively important or comparatively unimportant from the point of view of a particular theory''~\citep[p.597]{bogosian2017implementation}.
However, since it is difficult to select representative actions \citep[cf.][p.102]{macaskill-bykvist-ord2020moral-uncertainty}, we treat $\mathcal{G}$ as $\mathcal{A}$ in our experiment. This way of normalization relativizes the choiceworthiness scores to $\mathcal{A}$.

\paragraph{Step 4: Aggregation} 
Finally, we obtain expected choiceworthiness by following equation:
\begin{equation}
    CW^{E}\left( A \right) = \mathop \sum \limits_{i = 1}^{n} CW_{i}^{N} \left( A \right)C\left( {T_{i} } \right)
\end{equation}
The decision maker selects the action $A$ which maximizes $CW^E(A)$.

\section{Why is MEC preferable in Moral AI?}
As described above, MEC algorithm aggregates the output of models based on each normative ethical theory to output a final moral evaluation.
There are at least three reasons why this algorithm is desirable in Moral AI.

First, if one creates models based on theories, one can, in principle, evaluate the output of each theory-based model since there is a correct answer relative to the theory for every question. For example, when creating a utilitarian model, the evaluation of the utilitarian model can be evaluated by whether the model's output is appropriate from a utilitarian point of view. However, in the case of a model based on commonsense morality, not theory, such as Delphi, it is difficult to evaluate the model because people sometimes have differing opinions about moral problems such as moral dilemmas or political issues.

Second, because MEC is a general framework, it can be used for the models developed in this paper and other models that may be proposed or developed by others in the future.
Although we have only used three theories in this experiment, we can use existing models such as Delphi and future proposed models to produce moral evaluations.

Third, the models aggregated in the MEC need not necessarily be theory-based. For example, if one wants to reflect cultural diversity, one can create a Delphi-like model for each culture, and MEC can aggregate the output of each model. Of course, reflecting the commonsense morality of each culture without being based on ethical theory makes evaluation difficult, as noted in the first point, but the ability to reflect cultural diversity in this way is another advantage of MEC being a general framework. Although we did not use a model that reflects cultural diversity in our experiments, we plan to develop it in the future.

We did not use a model reflecting cultural diversity in this experiment, but plan to develop one in the future.

\section{Experiment}
\subsection{Implementation}
To calculate expected choiceworthiness, we need AI models based on ethical theories. 
For this purpose, we fine-tune DeBERTa-v3$_{\rm large}$~\citep{he2021debertav3}\footnote{\url{https://huggingface.co/microsoft/deberta-v3-large}}\footnote{We do not have enough computational resources to fine-tune large models such as T5$_{11B}$. We use DeBERTa-v3$_{\rm large}$ because it is one of the best performing models.} on datasets included in ETHICS~\citep{hendrycks2021aligning-AI-human-values}. 
DeBERTa-v3 is an ELECTRA-style pretrained model. 
ELECTRA~\citep{clark2020electra} is a model pre-trained through Replaced Token Detection (RTD). RTD is a model training method where some tokens in the original sentence are masked, Generator (Masked Language Model~\citep[cf.][]{devlin-etal-2019-bert}) fills the masked tokens with words using, and Detector detects the words filled in the mask. 
\citet{he2021debertav3} improved the RTD by Gradient-Disentangled Embedding Sharing. While the gradient is shared between Generator and Detector during training in ELECTRA, the gradient is split between Generator and Detector in DeBERTa-v3. 

For fine-tuning DeBERTa-v3, we use three datasets: ``utilitarianism'', ``deontology'' and ``virtue''. These theories are the most endorsed theories in normative ethics \citep{PhilPapersSurvey2020}\footnote{\url{https://survey2020.philpeople.org/survey/results/4890}}\footnote{PhilPapers Survey did not use ``utilitarianism'' but ``consequentialism''. Utilitarianism is a particular type of consequentialism, therefore we can use the ``utilitarianism'' dataset as a consequentialist dataset.}. 
We set all $C(T_i)$ to 1. \citet{bogosian2017implementation} suggested some ways of assigning $C(T_i)$, one of which is to assign the philosopher's endorsement rate. According to PhilPapers Survey \citep{PhilPapersSurvey2020}, the supporters of each theory ( consequentialism, deontology, and virtue ethics) are roughly equal (32\%, 31\%, and 37\%, respectively). Therefore we assign hypothetically equal values in this experiment. 
Hyperparameters are shown on Table \ref{tab:hyperparameters}.

\begin{table}[t]
    \centering
    \begin{tabular}{cc}\toprule
     Hyperparameter & \\\midrule
       learning rate  & $1 \times 10^{-5}$, $2 \times 10^{-5}$, $3 \times 10^{-5}$ \\
       batch size  & 64\\
       epoch & 4 \\
       optimizer & AdamW\\
       warmup ratio & 0.1\\\bottomrule
    \end{tabular}
    \caption{Hyperparameters. We use AdamW~\citep{loshchilov2019decoupled-adamW} as an optimizer with default hyperparameters in Pytorch~\citep{paszke2019_pytorch} \protect \footnotemark.}
    \label{tab:hyperparameters}
\end{table}
\footnotetext{\url{https://pytorch.org/docs/stable/generated/torch.optim.AdamW.html}}

There are two problems with using ETHICS dataset.
First, this dataset was created by nonspecialist crowdworkers. 
Second, this dataset was not annotated directly according to each normative theory. For example, according to utilitarianism, an action is right if and only if the action maximizes the total well-being of all sentient beings affected by the action. However, \citet{hendrycks2021aligning-AI-human-values} asked crowdworkers to evaluate the well-being of the person who is presented in the given sentence, not all sentient beings.
Hence, this dataset is not perfectly based on moral theories.
Nevertheless, we use ETHICS in our experiment because it is the only dataset created based on ethical theories.
As already mentioned, an advantage of our approach is that such a dataset and AI models can be refined based on moral theories. We plan to develop a more theory-informed dataset than ETHICS in the future.

In using each model for MEC, the following input-output structure is used. 
First, the utilitarian model outputs a scalar value and treats it directly as a choiceworthiness score, following the construct of \citet{hendrycks2021aligning-AI-human-values}.
Next, for the deontology model, we use the form ``I am a human [SEP] $a_s$'' as input, following the input form of the Role subtask, and treat its output value (the probability that the $a_s$ is permissible) as the choiceworthiness score, where $a_s$ denotes an action statement under consideration.
This input form is intended to allow the model to determine whether an action is morally permissible or impermissible as a human being.
Finally, for the virtue ethics model, we first list all character trait terms in the ``virtue'' train set and assign sentiment by SenticNet~\citep{cambria-etal-2022-senticnet7}. Terms not included in SenticNet are excluded.
As a result, we collected 695 character trait terms and their sentiment. 
Let $v_t$ be the term of a character trait, the input format is ``$a_s$ [SEP] $v_t$'', and the sentiment of $v_t$ with the highest probability is treated as the choiceworthiness score of $a_s$.

\subsection{Evaluation}
We assess the performance of our model through a evaluation process, which involves two distinct methods

\subsubsection{Experiment 1: Evaluate the performance and generalizability of each model using the ETHICS dataset}
First, we assess the results for the test set of each sub-dataset, i.e., ``utilitarianism'', ``deontology'', and ``virtue''.
We also use ``commonsense'' in ETHICS to evaluate generalizability of our models. 
We expect the accuracy of MEC in this dataset to be superior to the accuracy of each model because of aggregation.
In the case of the ``commonsense'' dataset, we set the threshold of the utilitarianism model for classification using 1,000 samples train data of the ``commonsense'' dataset. For the other two models (the deontology and the virtue ethics model), the output is positive if it is greater than zero and negative otherwise.

\begin{table*}[t]
    \centering
    \begin{tabular}{cccc}
        \toprule
       Model  &  Deontology & Virtue 
& Utilitarianism \\\midrule
   Random Baseline	& 6.3 / 6.3 & 8.2 / 8.2 & 50.0 / 50.0 \\
   BERT$_{\rm large}$ & 44.2 / 13.6 & 40.6 / 13.5 & 74.6 / 49.1 \\
     RoBERTa$_{\rm large}$ & 60.3 / 30.8 & 53.0 / 25.5 & 79.5 / 62.9 \\
    ALBERT$_{\rm xxlarge}$ & 64.1 / 37.2 & 64.1 / 37.8 & 81.9 / 67.4 \\
    T5$_{\rm 11B}$* & 16.9 / 11.0 & 1.6 / 0.8 & 82.8 / 70.4 \\
    Delphi* & 49.6 / 31.0 & 29.5 / 18.2 & \textbf{84.9} / \textbf{76.0} \\\midrule
     DeBERTa-v3$_{\rm large}$  & \textbf{78.0} / \textbf{59.4} & \textbf{76.3} / \textbf{50.9}   &  81.6 / 73.6\\\bottomrule
    \end{tabular}
    
    \caption{The results (\textbf{Test / Hard Test}) for the test set of ETHICS dataset~\citep{hendrycks2021aligning-AI-human-values}. The scores of BERT$_{\rm large}$~\citep{devlin-etal-2019-bert}, RoBERTa$_{\rm large}$~\citep{liu2019roberta} and ALBERT$_{\rm xxlarge}$~\citep{Lan2020ALBERT} are reported by \citet{hendrycks2021aligning-AI-human-values}, scores of T5$_{\rm 11B}$~\citep{Raffel-2020-t5} and Delphi are reported by \citet{Jiang-2021-can-machine-learn-morality-delphi}. The best scores are shown in \textbf{bold font}. * T5$_{\rm 11B}$ and Delphi are fine-tuned with only 100 sampled training instances.}
    \label{tab:result-ethics}
\end{table*}

\subsubsection{Experiment 2: Comparison with Delphi by asking Ph.D. students}
Second, we compare our model with Delphi~\citep{Jiang-2021-can-machine-learn-morality-delphi} by asking three Ph.D. students majoring in philosophy~\footnote{We asked, in English, three Japanese Ph.D. students who are fluent in English. There might be some minor language problems, but we do not think they significantly influence the results.}.

There are two kinds of evaluation: each output and the overall. First, we asses the output of our models using 40 sampled data (20 pairs) of test sets of ``commonsense'' in ETHICS dataset. We ask the annotators if the outputs of each of the models were properly theory-based, respectively.
We also asked whether the Delphi's outputs and the aggregated output (i.e. MEC's output) were consistent with the annotators' moral judgments.

Second, in overall evaluation, we ask the Ph.D. students to compare Delphi and MEC model with three metrics and the reasons:
\begin{enumerate}
    \item Which is preferred: one answer output to one question (like Delphi) or a comparative ranking for options (like MEC)? Why?
        \begin{enumerate}
            \item The former is more preferable than the latter.
            \item The latter is more preferable than the former.
            \item Both are preferable.
            \item Both are not preferable.
        \end{enumerate}
    \item When a \textit{non-expert} in ethics were to use Delphi or MEC as an AI advisor in moral deliberation, which model would be helpful? Why?
        \begin{enumerate}
            \item Delphi is more helpful than MEC.
            \item MEC is more helpful than Delphi.
            \item Both are helpful.
            \item Both are not helpful.
        \end{enumerate}
    \item When an \textit{expert} in ethics were to use Delphi or MEC as an AI advisor in moral deliberation, which model would be helpful? Why?
        \begin{enumerate}
            \item Delphi is more helpful than MEC.
            \item MEC is more helpful than Delphi.
            \item Both are helpful.
            \item Both are not helpful.
        \end{enumerate}
\end{enumerate}

We ask people to evaluate these models as AI advisors for two reasons. 
First, when people use these models, they ask the AI for opinions about ethics and use it as a decision-making tools~\citep{anderson2006medethex, anderson-anderson-2018-geneth}.
Second, there may be a kind of explanation of the model's output by showing users not only the aggregated results of the MEC outputs but also the outputs of each model on which the aggregation is based.
If this explains the model's output, it should be more useful in moral deliberation. In contrast, since Delphi does not know the reason for its outputs, we believe that the outputs from MEC are superior in this respect, and we use this metric to confirm this.
Also, for these reasons, MEC may be helpful to non-expert but not to experts in ethics. Therefore, we examine this hypothesis through questions 2 and 3.

\begin{table}[t]
    \centering
    \begin{tabular}{cc}\toprule
     Model   &  Commonsense (acc)\\\midrule
     Random Baseline & 50.0\\
      Utilitarianism Model  & 76.5 \\
      Deontology Model  & 71.5 \\
      Virtue Ethics Model &  77.1 \\
      MEC & \textbf{82.3} \\\bottomrule
    \end{tabular}
    \caption{The results for the test set (only short sentence) of ``commonsense'' dataset in ETHICS~\citep{hendrycks2021aligning-AI-human-values}. The best score is shown in \textbf{bold font}.}
    \label{tab:result-commonsense}
\end{table}

\section{Results}
\subsection{Experiment 1: performance and generalizability of each model using the ETHICS dataset}

\begin{table*}[ht]
    \centering
    \begin{tabular}{p{7cm}c}\toprule
       Question  &  Answer\\\midrule
    Which is preferred, outputting one answer to one question (like Delphi) or a comparative ranking for options (like MEC)?
    & \begin{tabular}{c}
        The former (like Delphi) is more preferable than the latter. (2/3)  \\
         The latter (like MEC) is more preferable than the former. (1/3)
    \end{tabular}  \\\midrule
    When a \textit{non-expert} in ethics were to use Delphi or MEC as an AI advisor in moral deliberation, which model would be helpful? & 
    \begin{tabular}{c}
       MEC is more helpful than Delphi. (1/3)   \\
    Both are not helpful. (2/3)
    \end{tabular}
 \\\midrule
    When an \textit{expert} in ethics were to use Delphi or MEC as an AI advisor in moral deliberation, which model would be helpful? & 
    Both are not helpful. (3/3)\\ \bottomrule
    \end{tabular}
    \caption{The results for each response in Experiment 2. Numbers in parentheses indicate the number of annotators who chose that response.}
    \label{tab:comparison_delphi_phd}
\end{table*}

\begin{table}[t]
    \centering
    \begin{tabular}{cc}\toprule
       Delphi  &  MEC\\\midrule
       13/16 (81\%)  & \textbf{14/16 (88\%)}\\\bottomrule
    \end{tabular}
    \caption{The results for the paired test set of the ETHICS dataset obtained by asking Ph.D. students. The reason for the 16 instead of 20 evaluations is that some instances were not considered evaluable due to a lack of context.}
    \label{tab:comparison_delphi_mec}
\end{table}

\begin{table}[t]
    \centering
    \begin{tabular}{ccc}\toprule
     Utilitarianism    &  Deontology & Virtue Ethics \\\midrule
       12/12 (100\%)  &  13/15 (87\%) &  17/17 (100\%)\\ \bottomrule
    \end{tabular}
    \caption{The results for the paired test set of the ETHICS dataset were obtained by asking Ph.D. students. 
    The reason for number of evaluations smaller than 20 is that some instances were not considered evaluable due to a lack of context.}
    \label{tab:acc_util_deon_virtue_phd}
\end{table}

We show the results of ETHICS dataset as the test set in Tables \ref{tab:result-ethics} and \ref{tab:result-commonsense}.
Except for utilitarianism dataset, DeBERTa results are better than BERT, RoBERTa, and ALBERT results, which are fine-tuned with all train data. It also outperforms the T5 and Delphi, which are fine-tuned with 100 sampled train data.
The higher accuracy for the other models on the utilitarian dataset is likely because this task was designed as a binary classification task.

Regarding the results for the commonsense dataset, since MEC is an ensemble model of each theory-based model, the results are better than the other three models as expected.
In addition, each model outperforms the random baseline even though it was not fine-tuned on the commonsense dataset. These results indicate that the verdict based on each theory correlates to some extent with the commonsense morality verdict.
This correlation suggests generalizability from learning on each theory to commonsense morality.
However, the accuracy of each theory-based model on this dataset is relatively low, and we will examine how it could achieve higher performance in the future.

\subsection{Experiment 2: Comparison with Delphi by asking Ph.D. students}

In Table \ref{tab:comparison_delphi_mec} we show the results of the questioning  Ph.D. students majoring in philosophy . 
We asked three annotators to evaluate 20 paired sentences, but they were not able to evaluate four paired sentences because of a lack of context. Therefore, we are showing the results for 16 pairs. Delphi and MEC showed little difference in the results. Delphi, trained on the ``commonsense'' dataset, was expected to be more accurate than MEC, but this was not the case.  

We also show the results based on each theory (see Table \ref{tab:acc_util_deon_virtue_phd}) only for the paired sentences answered by all three students because some paired sentences were not evaluated due to a lack of context. 
All three models used in MEC had a high percentage of correct answers.

Next, Table \ref{tab:comparison_delphi_phd} shows the results of the overall evaluation of Delphi and MEC.
Regarding the first question, most annotators preferred to give one answer to one question, as in Delphi. 
The reasons given were that people only care about the absolute evaluation (non-comparative evaluation) of the alternatives and that it does not work well when the evaluations are in the same order.
On the other hand, one annotator answered that it is preferable to provide a ranking of options, as in the MEC. The reason is that if only a single answer is output and different from the expectation, the impression is that it cannot be used as a reference. However, if it is output as a ranking, it is likely to generate a certain degree of acceptance.

For the second question, there was only one answer that MEC is more helpful than Delphi, and the reason was ``because referencing multiple positions seems more plausible.
The remaining answers were that both are not helpful because reasons are essential in moral deliberation, but neither model provides any.

Concerning the third question, all annotators indicated that both systems are not helpful, and the reason given was that neither model provided a reason. In the case of the experts, it seems to be more preferable to think for themselves.

\section{Discussion}
\subsection{On the performance of MEC and theory-based models}
As seen from the results of Experiment 1 (Table \ref{tab:result-ethics}), DeBERTa-v3 yielded superior performance compared to the baseline model, T5 and Delphi. In addition, Experiment 2 (Table \ref{tab:acc_util_deon_virtue_phd}) also showed that DeBERTa-v3 is reliable enough to be used in this experiment, as it produces appropriate results based on each theory. However, the evaluation of the ETHICS dataset on the Hard Test set is still low and needs further improvement for a practical use.

All models outperform the random baseline for the results for the MEC ``commonsense'' dataset in Experiment 1, suggesting that each theory generally reflects commonsense morality.
Although it is said that utilitarianism is generally counterintuitive, this theory reflects some of our commonsense morality. The results suggest utilitarianism is not entirely uncorrelated with commonsense morality. 
Moreover, other theories may also make counterintuitive judgments in some cases (e.g. Kantian obligation to never lie).
The advantage of being theory-based is that theory-based judgment does not follow directly from our commonsense morality. Therefore, it is desirable that the model based on each theory does not perfectly agree with our commonsense moral evaluations. 
Nevertheless, a theory-based model must correlate with common sense morality because if it is too different from common sense morality, people will not want to use such a model. The extent to which commonsense morality and the evaluation of each theory should be correlated, and the extent to which they should not be correlated, is a controversial problem.
Furthermore, we will examine situations where there is a discrepancy between commonsense morality judgments and theoretical judgments since the ``commonsense'' dataset does not cover controversial or politically divisive topics.

\subsection{Comparing MEC with Delphi}
Annotators answered that models such as Delphi and MEC are not helpful in moral deliberation, both for experts and non-experts, because they do not provide reasons, i.e., they do not explain their choices.
This result rejects our original hypothesis that MEC is a kind of explanation because the model outputs the results of theory-based models.
It may not be an explanation unless the models also provide reasons for why the model on which each theory is based produces the output it does.
We can solve this problem by preparing a template for the output of each theory-based model. For example, a utilitarian model could have a template such as "Action A is more choiceworthy than action B because action A has higher utility than action B."
In addition, it would be desirable if it is possible to explain, for example, why action A has higher utility than action B~\citep[cf.][]{Bang2022TowardsAnswering-openend-ethical-questions}.
We will investigate what type of output format is appropriate in the future.

One of the annotators also stated that MEC could be more helpful depending on its use. For example, Delphi and MEC would promote moral deliberation if users reconsidered their judgment based on their output. In this case, MEC can provide aggregated outputs and the outputs of each theory-based model, respectively, promoting moral deliberation more than Delphi.
\citet{takeshita2023defence-moralAIenhancement_en} suggests a variety of possible uses for such Moral AI, some of which would assist users in moral deliberation and help them make more appropriate moral decisions than those not used.
In the future, we will investigate what outputs can better support users' moral deliberation.

Next, regarding the comparison between experts and non-experts, it was found that both MEC and Delphi were not helpful for experts, as we hypothesized. 
However, our hypothesis that MEC is helpful for non-experts was not supported because, as already mentioned, MEC does not provide reasons or explanations.
On the other hand, one of the annotators stated that MEC is more likely to guide the user's moral judgment more appropriately because it refers to multiple theories. 
This result was expected, because the output of the MEC algorithm is more appropriate than if it were based on only a single theory since it maximizes the expected choice worthiness.
Thus, in some cases, MEC may be useful to non-experts. We will explore in which cases MEC may be helpful to non-experts.

\section{Conclusion}
In this paper, we implemented and evaluated Maximizing Expected Choiceworthiness algorithm. This algorithm aggregates the output of models based on multiple normative theories to generate a morally correct output. Furthermore, this model produces appropriate outputs under moral uncertainty when the morally correct theory is unknown. Experimental results show that MEC is more compatible with commonsense morality than a single model and performs as well as Delphi, an existing method. However, this model does not provide enough reasons or explanations, and we plan to create models that provide more reasons or explanations in the future.

\section*{Limitations}
The dataset based on each theory included in the ETHICS dataset used in this study does not precisely match the evaluation based on each theory discussed in Section 5.1. Therefore, the model created in this study may not be ideally based on each theory.
Moreover, the scale of the experiments is small. In Experiment 2, where the model is evaluated by asking Ph.D. students, only a maximum of 16 pairs of sentences are used. Furthermore, we did not evaluate our model in complex cases such as moral dilemmas.

\section*{Ethical and Social Implications}
There is no guarantee that the output of the model implemented in this study is morally correct. Continued refinement of this model will yield more appropriate outputs, but the current model is inadequate.
We also do not recommend that users rely on the output of our model (or its improved versions) to make decisions. Our model is only a decision support tool, not a substitute for user decision-making.

Our model can contribute to the AI safety. Moreover, as ethical theory developed and models based on them can be created, MEC algorithm will make AI behavior more ethically appropriate.

\section*{Acknowledgement}
This work was supported by JSPS KAKENHI Grant Number JP22J21160.
We would like to thank the anonymous reviewers for their valuable comments.

\bibliographystyle{named}
\bibliography{mybib}

\begin{thebibliography}{}

\bibitem[\protect\citeauthoryear{Allen \bgroup \em et al.\egroup
  }{2005}]{allen2005artificial-morality}
Colin Allen, Iva Smit, and Wendell Wallach.
\newblock Artificial morality: Top-down, bottom-up, and hybrid approaches.
\newblock {\em Ethics and information technology}, 7:149--155, 2005.

\bibitem[\protect\citeauthoryear{Anderson and
  Anderson}{2018}]{anderson-anderson-2018-geneth}
Michael Anderson and Susan~Leigh Anderson.
\newblock Gen{E}th: A general ethical dilemma analyzer.
\newblock {\em Paladyn, Journal of Behavioral Robotics}, 9(1):337--357, 2018.

\bibitem[\protect\citeauthoryear{Anderson \bgroup \em et al.\egroup
  }{2006}]{anderson2006medethex}
Michael Anderson, Susan~Leigh Anderson, and Chris Armen.
\newblock Med{E}th{E}x: a prototype medical ethics advisor.
\newblock In {\em Proceedings of the {N}ational {C}onference on {A}rtificial
  {I}ntelligence}, volume~21, page 1759. Menlo Park, CA; Cambridge, MA; London;
  AAAI Press; MIT Press; 1999, 2006.

\bibitem[\protect\citeauthoryear{Awad \bgroup \em et al.\egroup
  }{2018}]{awad2018moral-machine-experiment}
Edmond Awad, Sohan Dsouza, Richard Kim, Jonathan Schulz, Joseph Henrich, Azim
  Shariff, Jean-Fran{\c{c}}ois Bonnefon, and Iyad Rahwan.
\newblock The moral machine experiment.
\newblock {\em Nature}, 563(7729):59--64, 2018.

\bibitem[\protect\citeauthoryear{Bang \bgroup \em et al.\egroup
  }{2022}]{Bang2022TowardsAnswering-openend-ethical-questions}
Yejin Bang, Nayeon Lee, Tiezheng Yu, Leila Khalatbari, Yan Xu, Samuel
  Cahyawijaya, Dan Su, Bryan Wilie, Romain Barraud, Elham~J. Barezi, Andrea
  Madotto, Hayden Kee, and Pascale Fung.
\newblock Towards answering open-ended ethical quandary questions.
\newblock {\em arXiv preprint arXiv:2205.05989}, 2022.

\bibitem[\protect\citeauthoryear{Bogosian}{2017}]{bogosian2017implementation}
Kyle Bogosian.
\newblock Implementation of moral uncertainty in intelligent machines.
\newblock {\em Minds and Machines}, 27:591--608, 2017.

\bibitem[\protect\citeauthoryear{Bommasani \bgroup \em et al.\egroup
  }{2021}]{bommasani2021opportunities-risks-foundation-models}
Rishi Bommasani, Drew~A Hudson, Ehsan Adeli, Russ Altman, Simran Arora, Sydney
  von Arx, Michael~S Bernstein, Jeannette Bohg, Antoine Bosselut, Emma
  Brunskill, et~al.
\newblock On the opportunities and risks of foundation models.
\newblock {\em arXiv preprint arXiv:2108.07258}, 2021.

\bibitem[\protect\citeauthoryear{Bourget and
  Chalmers}{ms}]{PhilPapersSurvey2020}
David Bourget and David Chalmers.
\newblock Philosophers on philosophy: The philpapers 2020 survey, ms.

\bibitem[\protect\citeauthoryear{Braun \bgroup \em et al.\egroup
  }{2021}]{Braune-2021-primer-ethics-ai-decision-support-clinic}
Matthias Braun, Patrik Hummel, Susanne Beck, and Peter Dabrock.
\newblock Primer on an ethics of {AI}-based decision support systems in the
  clinic.
\newblock {\em Journal of Medical Ethics}, 47(12):e3--e3, 2021.

\bibitem[\protect\citeauthoryear{Brown \bgroup \em et al.\egroup
  }{2020}]{brown2020language-few-shot-leaner-gpt-3}
Tom Brown, Benjamin Mann, Nick Ryder, Melanie Subbiah, Jared~D Kaplan, Prafulla
  Dhariwal, Arvind Neelakantan, Pranav Shyam, Girish Sastry, Amanda Askell,
  et~al.
\newblock Language models are few-shot learners.
\newblock {\em Advances in neural information processing systems},
  33:1877--1901, 2020.

\bibitem[\protect\citeauthoryear{Cambria \bgroup \em et al.\egroup
  }{2022}]{cambria-etal-2022-senticnet7}
Erik Cambria, Qian Liu, Sergio Decherchi, Frank Xing, and Kenneth Kwok.
\newblock {S}entic{N}et 7: A commonsense-based neurosymbolic {AI} framework for
  explainable sentiment analysis.
\newblock In {\em Proceedings of the Thirteenth Language Resources and
  Evaluation Conference}, pages 3829--3839, Marseille, France, June 2022.
  European Language Resources Association.

\bibitem[\protect\citeauthoryear{Clark \bgroup \em et al.\egroup
  }{2020}]{clark2020electra}
Kevin Clark, Minh-Thang Luong, Quoc~V. Le, and Christopher~D. Manning.
\newblock Electra: Pre-training text encoders as discriminators rather than
  generators.
\newblock In {\em International Conference on Learning Representations}, 2020.

\bibitem[\protect\citeauthoryear{Devlin \bgroup \em et al.\egroup
  }{2019}]{devlin-etal-2019-bert}
Jacob Devlin, Ming-Wei Chang, Kenton Lee, and Kristina Toutanova.
\newblock {BERT}: Pre-training of deep bidirectional transformers for language
  understanding.
\newblock In {\em Proceedings of the 2019 Conference of the North {A}merican
  Chapter of the Association for Computational Linguistics: Human Language
  Technologies, Volume 1 (Long and Short Papers)}, pages 4171--4186.
  Association for Computational Linguistics, 2019.

\bibitem[\protect\citeauthoryear{Emelin \bgroup \em et al.\egroup
  }{2021}]{emelin-etal-2021-moral-stories}
Denis Emelin, Ronan Le~Bras, Jena~D. Hwang, Maxwell Forbes, and Yejin Choi.
\newblock Moral {S}tories: Situated reasoning about norms, intents, actions,
  and their consequences.
\newblock In {\em Proceedings of the 2021 Conference on Empirical Methods in
  Natural Language Processing}, pages 698--718. Association for Computational
  Linguistics, 2021.

\bibitem[\protect\citeauthoryear{Forbes \bgroup \em et al.\egroup
  }{2020}]{forbes-etal-2020-social-chemistry-101}
Maxwell Forbes, Jena~D. Hwang, Vered Shwartz, Maarten Sap, and Yejin Choi.
\newblock Social chemistry 101: Learning to reason about social and moral
  norms.
\newblock In {\em Proceedings of the 2020 Conference on Empirical Methods in
  Natural Language Processing (EMNLP)}, pages 653--670. Association for
  Computational Linguistics, 2020.

\bibitem[\protect\citeauthoryear{Ganguli \bgroup \em et al.\egroup
  }{2022}]{ganguli2022red-teaming-languagemodel-method-behavior}
Deep Ganguli, Liane Lovitt, Jackson Kernion, Amanda Askell, Yuntao Bai, Saurav
  Kadavath, Ben Mann, Ethan Perez, Nicholas Schiefer, Kamal Ndousse, et~al.
\newblock Red teaming language models to reduce harms: Methods, scaling
  behaviors, and lessons learned.
\newblock {\em arXiv preprint arXiv:2209.07858}, 2022.

\bibitem[\protect\citeauthoryear{Gehman \bgroup \em et al.\egroup
  }{2020}]{gehman-etal-2020-realtoxicityprompts}
Samuel Gehman, Suchin Gururangan, Maarten Sap, Yejin Choi, and Noah~A. Smith.
\newblock {R}eal{T}oxicity{P}rompts: Evaluating neural toxic degeneration in
  language models.
\newblock In {\em Findings of the Association for Computational Linguistics:
  EMNLP 2020}, pages 3356--3369, Online, November 2020. Association for
  Computational Linguistics.

\bibitem[\protect\citeauthoryear{He \bgroup \em et al.\egroup
  }{2021}]{he2021debertav3}
Pengcheng He, Jianfeng Gao, and Weizhu Chen.
\newblock Debertav3: Improving deberta using {ELECTRA}-style pre-training with
  gradient-disentangled embedding sharing.
\newblock {\em arXiv preprint arXiv:2111.09543}, 2021.

\bibitem[\protect\citeauthoryear{Hendrycks \bgroup \em et al.\egroup
  }{2021}]{hendrycks2021aligning-AI-human-values}
Dan Hendrycks, Collin Burns, Steven Basart, Andrew Critch, Jerry Li, Dawn Song,
  and Jacob Steinhardt.
\newblock Aligning {AI} with shared human values.
\newblock In {\em International Conference on Learning Representations}, 2021.

\bibitem[\protect\citeauthoryear{Jiang \bgroup \em et al.\egroup
  }{2021}]{Jiang-2021-can-machine-learn-morality-delphi}
Liwei Jiang, Jena~D. Hwang, Chandra Bhagavatula, Ronan~Le Bras, Jenny Liang,
  Jesse Dodge, Keisuke Sakaguchi, Maxwell Forbes, Jon Borchardt, Saadia
  Gabriel, Yulia Tsvetkov, Oren Etzioni, Maarten Sap, Regina Rini, and Yejin
  Choi.
\newblock Can machines learn morality? the {D}elphi experiment.
\newblock {\em arXiv preprint arXiv:2110.07574}, 2021.

\bibitem[\protect\citeauthoryear{Jin \bgroup \em et al.\egroup
  }{2022}]{Jin-2022-when-to-make-exceptions-moral-exceptqa}
Zhijing Jin, Sydney Levine, Fernando Gonzalez~Adauto, Ojasv Kamal, Maarten Sap,
  Mrinmaya Sachan, Rada Mihalcea, Josh Tenenbaum, and Bernhard Sch\"{o}lkopf.
\newblock When to make exceptions: Exploring language models as accounts of
  human moral judgment.
\newblock In S.~Koyejo, S.~Mohamed, A.~Agarwal, D.~Belgrave, K.~Cho, and A.~Oh,
  editors, {\em Advances in Neural Information Processing Systems}, volume~35,
  pages 28458--28473. Curran Associates, Inc., 2022.

\bibitem[\protect\citeauthoryear{Lan \bgroup \em et al.\egroup
  }{2020}]{Lan2020ALBERT}
Zhenzhong Lan, Mingda Chen, Sebastian Goodman, Kevin Gimpel, Piyush Sharma, and
  Radu Soricut.
\newblock A{LBERT}: A lite {BERT} for self-supervised learning of language
  representations.
\newblock In {\em International Conference on Learning Representations}, 2020.

\bibitem[\protect\citeauthoryear{Liu \bgroup \em et al.\egroup
  }{2019}]{liu2019roberta}
Yinhan Liu, Myle Ott, Naman Goyal, Jingfei Du, Mandar Joshi, Danqi Chen, Omer
  Levy, Mike Lewis, Luke Zettlemoyer, and Veselin Stoyanov.
\newblock Ro{BERT}a: A robustly optimized {BERT} pretraining approach.
\newblock {\em Computing Research Repository}, arXiv:1907.11692, 2019.

\bibitem[\protect\citeauthoryear{Loshchilov and
  Hutter}{2019}]{loshchilov2019decoupled-adamW}
Ilya Loshchilov and Frank Hutter.
\newblock Decoupled weight decay regularization.
\newblock In {\em International Conference on Learning Representations}, 2019.

\bibitem[\protect\citeauthoryear{Lourie \bgroup \em et al.\egroup
  }{2021}]{lourie2021unicorn-rainbow}
Nicholas Lourie, Ronan Le~Bras, Chandra Bhagavatula, and Yejin Choi.
\newblock Unicorn on rainbow: A universal commonsense reasoning model on a new
  multitask benchmark.
\newblock In {\em Proceedings of the AAAI Conference on Artificial
  Intelligence}, volume~35, pages 13480--13488, 2021.

\bibitem[\protect\citeauthoryear{MacAskill \bgroup \em et al.\egroup
  }{2020}]{macaskill-bykvist-ord2020moral-uncertainty}
Michael MacAskill, Krister Bykvist, and Toby Ord.
\newblock {\em Moral uncertainty}.
\newblock Oxford University Press, 2020.

\bibitem[\protect\citeauthoryear{Paszke \bgroup \em et al.\egroup
  }{2019}]{paszke2019_pytorch}
Adam Paszke, Sam Gross, Francisco Massa, Adam Lerer, James Bradbury, Gregory
  Chanan, Trevor Killeen, Zeming Lin, Natalia Gimelshein, Luca Antiga, Alban
  Desmaison, Andreas Kopf, Edward Yang, Zachary DeVito, Martin Raison, Alykhan
  Tejani, Sasank Chilamkurthy, Benoit Steiner, Lu~Fang, Junjie Bai, and Soumith
  Chintala.
\newblock Pytorch: An imperative style, high-performance deep learning library.
\newblock In H.~Wallach, H.~Larochelle, A.~Beygelzimer, F.~d\textquotesingle
  Alch\'{e}-Buc, E.~Fox, and R.~Garnett, editors, {\em Advances in Neural
  Information Processing Systems}, volume~32. Curran Associates, Inc., 2019.

\bibitem[\protect\citeauthoryear{Raffel \bgroup \em et al.\egroup
  }{2020}]{Raffel-2020-t5}
Colin Raffel, Noam Shazeer, Adam Roberts, Katherine Lee, Sharan Narang, Michael
  Matena, Yanqi Zhou, Wei Li, and Peter~J. Liu.
\newblock Exploring the limits of transfer learning with a unified text-to-text
  transformer.
\newblock {\em Journal of Machine Learning Research}, 21(140):1--67, 2020.

\bibitem[\protect\citeauthoryear{Rzepka and Araki}{2017}]{Rzepka2017WhatPS}
Rafal Rzepka and Kenji Araki.
\newblock What people say? web-based casuistry for artificial morality
  experiments.
\newblock In {\em Artificial {G}eneral {I}ntelligence}, 2017.

\bibitem[\protect\citeauthoryear{Takeshita}{2023}]{takeshita2023defence-moralAIenhancement_en}
Masashi Takeshita.
\newblock A defense of moral {AI} enhancement (in {J}apansese).
\newblock {\em Applied Ethics}, 14:3--20, 2023.

\bibitem[\protect\citeauthoryear{Tolmeijer \bgroup \em et al.\egroup
  }{2020}]{tolmeijer2020implementations-machine-ethics}
Suzanne Tolmeijer, Markus Kneer, Cristina Sarasua, Markus Christen, and Abraham
  Bernstein.
\newblock Implementations in machine ethics: A survey.
\newblock {\em ACM Computing Surveys (CSUR)}, 53(6):1--38, 2020.

\bibitem[\protect\citeauthoryear{Wallach and
  Allen}{2008}]{wallach-allen2008moral-machines}
Wendell Wallach and Colin Allen.
\newblock {\em Moral machines: Teaching robots right from wrong}.
\newblock Oxford University Press, 2008.

\end{thebibliography}

\end{document}